\documentclass[a4paper,12pt,twoside]{article}
\usepackage{graphicx,wl4-mydefs}
\begin{document}
\title{\bf\Large Testing for Mathematical Lineation in\\
Jim Crace's {\it Quarantine}\\
and T.~S.~Eliot's {\it Four Quartets}}
\author{{\large John Constable}\\
{\tt jbc12@cam.ac.uk}\\
{\it Magdalene College, Cambridge CB3 0AG, United Kingdom}\\
and\\
Hideaki Aoyama\\
{\tt aoyama@phys.h.kyoto-u.ac.jp}\\
\it Faculty of Integrated Human Studies\\
\it Kyoto University, Kyoto 606-8501, Japan}
\date{}
\maketitle
\begin{abstract}
The mathematical distinction between prose and verse may be detected in writings that are not apparently lineated, for example in T. S. Eliot's {\it Burnt Norton}, and Jim Crace's {\it Quarantine}. In this paper we offer comments on appropriate statistical methods for such work, and also on the nature of formal innovation in these two texts. Additional remarks are made on the roots of lineation as a metrical form, and on the prose-verse continuum. 
\end{abstract}

\section{Introduction}
In Aoyama and Constable (1999) and Constable and Aoyama (1999)
we have developed a technique, the $Q_n$  distribution, for detecting 
a mathematical distinction between prose and 
isometrically lineated text (verse) in English. For prose the $Q_n$  
distribution is flat, for isometrically lineated verse it is peaked
at the value of $n$ which equals the number of syllables in the normative 
line. Our major focus in that work was on
the theoretical value of such a distinction, while in this paper we 
will turn to look more closely at a point raised before only in passing
(see Constable and Aoyama (1999:515)), namely the possibility 
that the features we
have identified might be employed as a `test' or diagnostic for 
lineation, enabling an analyst to detect the presence of mathematically
lineated text where it is not visually represented on the page (syllabic verse
printed as prose for example)  or is concealed by the layout on the page, 
as for example when a sonnet 
is embedded in prose. We cautioned against the belief that our 
procedure would enable the detection of small passages of verse embedded in prose on the grounds that the effects of such lineation on the $Q_n$ 
distribution would be so small
that any peaks resulting would be swamped by random fluctuations in 
the rest of the distribution.
However, we suggested that for longer texts the case might be different, 
and our aim here is to examine
two works, T. S. Eliot's poem {\it Four Quartets} (1936-1943)  and Jim Crace's novel {\it Quarantine} (1997)
with a view to assessing the
reliability of the procedure, and developing statistical techniques for 
determining the strength of inferences
based on the computation of $Q_n$ distributions.

In concluding we offer further comments on the experimental goals of these two texts, the roots of lineation as a metrical form, and the prose-verse continuum.

\section{Mathematical Preliminaries}
The essential feature of the syllabic structure of 
English prose found by Aoyama and Constable (1999) is  
{\it random segmentation}; by which we mean that the
probability of having a word boundary after any syllable 
is constant.  
Mathematically, this implies that the probability of
the syllable length of the $k$-th word, $S_k$, being equal
to $S$ is given by a geometric distribution;
\begin{equation}
P\{S_k=S\}=q(1-q)^{S-1},
\label{eqn:geom}
\end{equation}
where $q$ is the probability of the occurrence of a
word boundary after any syllable, this latter probability being a parameter
which varies between data sets.
The reader might note that the right hand side of the probability
(\ref{eqn:geom}) is independent from $k$, that is, 
the probability distribution does not depend on where the
word is located in the whole article.

This property of random segmentation leads to the finding that
the expectation value of the number of syllables
of the $k$-th word is independent from $k$;
\begin{equation}
E\{S_k\}=s, 
\label{eqn:whitenoise1}
\end{equation}
where the mean syllable length $s$ is given by $s=1/q$, 
and the following results for the expectation value of 
the products of the syllable lengths of the $k$-th word and the $k'$-th word:
\begin{equation}
E\{(S_k-E\{S_k\})(S_{k'}-E\{S_{k'}\})\}=\delta_{k,k'}\Delta,
\label{eqn:whitenoise2}
\end{equation}
where the variance $\Delta$ is given by $\Delta=(1-q)/q^2$.

The finding noted above was made mainly through studies of
the $Q_n$ distribution, supplemented with studies of the 
correlations between word lengths.
In Constable and Aoyama (1999), we found by examining the $Q_n$ distributions that isometrically lineated verse shows systematic deviation from the above properties.
In investigating the properties of {\it Quarantine} and
{\it Four Quartets}, we find it useful to utilize two other
standard statistical tools, Fourier analysis and 
Correlation functions,
in addition to the $Q_n$ distribution.
In the following, we will define these three quantities 
to prepare for the analysis.

\subsection{Fourier analysis}
The Fourier component $\tilde{S}_m$ is defined by the following 
equation:
\begin{equation}
S_k=\frac1{\sqrt{K}}\sum_{m=0}^{K-1} \tilde{S}_m e^{2\pi i mk/K},
\label{fouriercom}
\end{equation}
where $K$ is the total number of words in the data set.
The Fourier component can be directly calculated by 
the following inversion formula;
\begin{equation}
\tilde{S}_m=\frac1{\sqrt{K}}\sum_{k=1}^K S_k e^{-2\pi i mk/K}.
\end{equation}
These Fourier components satisfy the following relations:
\begin{eqnarray}
\tilde{S}_{m+K}&=&\tilde{S}_m,\\
\tilde{S}_{K-m}&=&\tilde{S}^*_m.\label{eqn:symm}
\end{eqnarray}

If the data set is randomly segmented, 
the expectation values for the Fourier coefficients $\tilde{S}_m$
satisfy the following equation:
\begin{equation}
E\{\tilde{S}_m\}=\sqrt{K} s\, \delta_{m,0}, \quad
E\{|\tilde{S}_m|^2\}=\Delta+\frac{\,s^2}{K}\,\delta_{m,0}, \quad
\end{equation}
which can be proved by using 
Eqs.(\ref{eqn:whitenoise1}) and (\ref{eqn:whitenoise2}).

The Fourier analysis is sensitive to any
periodic structure in the data; if there is a periodicity with
a period of $\ell$ in the sequence $\{S_1, S_2, \cdots, S_K\}$,
the coefficient $\tilde{S}_{K/\ell}$ (or its absolute value) will 
be large compared to the other Fourier coefficients.
The degree of predominance depends on the strength of the
periodicity: If the periodicity is weak the predominance of 
$\tilde{S}_\ell$ will be weak.

\subsection{The Correlation Function of Word Length}
We will  now turn to a consideration of the correlation function:
\begin{equation}
G_\ell\equiv
\frac{E\{(S_k-E\{S_k\})(S_{k+\ell}-E\{S_{k+\ell}\})\}}
{E\{(S_k-E\{S_k\})^2\}}.
\end{equation}
where the subscript of $S$ is defined by modulus $K$, i.e.,
$S_{K+1}\equiv S_1$, and so on. Since the value of $K$ is typically of 
the order $10^3 \gg 1$, this does not greatly affect the value of $G_\ell$.
From Eq.(\ref{eqn:whitenoise2}), it is evident that the
randomly segmented data leads to the following:
\begin{equation}
G_\ell=\delta_{k,k'}
\label{deltakk}
\end{equation}

\subsection{The $Q_n$ distribution}
The probability $Q_n$ is defined as the probability
that a sequence of adjacent words has
the total number of syllables $n$.
To be precise, let us define $L_{n,k}$ to be the number
of occurrences that $k$ sequential words have $n$ syllables in 
total.  From this definition, it is evident that the
following identity is satisfied:
\begin{equation}
\sum_{n=1}^\infty L_{n,k}=K.
\end{equation}
This is because since there are $K$ words in the data, there are $K$ 
sequences. 
The $Q_n$ distribution is defined by the following:
\begin{equation}
Q_n\equiv \frac1K \sum_{k=1}^n L_{n,k}.
\end{equation}
The upper limit of the sum in the above is induced by the property
that $L_{n,k}=0$ for $n<k$, which follows from the fact
that any English word is at least one-syllable long.

An alternative, but equivalent, definition
of $Q_n$ is that it is the probability that a word boundary occurs 
$n$-syllables after a word boundary. 
In this sense, $Q_n$ may be called the word-boundary correlation function.

\section{Jim Crace {\it Quarantine}}
\subsection{Fourier analysis}
\begin{figure}[ht]
\begin{center}
\includegraphics[width=6cm]{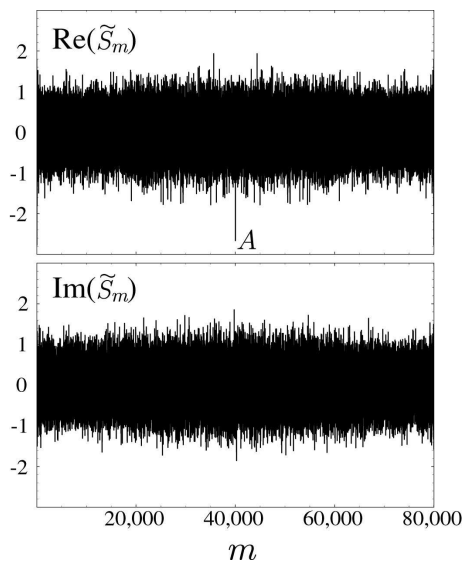}
\caption{The real part (upper plot) and the
imaginary part (lower plot) of the Fourier components
$\tilde{S}_m$ for {\it Quarantine} for $m=1$ to $80,009$.}
\label{fig:qfourier}
\end{center}
\end{figure}
The plot of the Fourier coefficients $\tilde{S}_m$
for {\it Quarantine} (with $K=80,010$ words)
is given in Fig.\ref{fig:qfourier}.
Due to the identity (\ref{eqn:symm}), the real part of
$\tilde{S}_m$ is symmetric around the middle-point $m=K/2$,
while the imaginary part of $\tilde{S}_m$ is anti-symmetric
around the middle-point.
 
\begin{figure}[ht]
\begin{center}
\includegraphics[width=11cm]{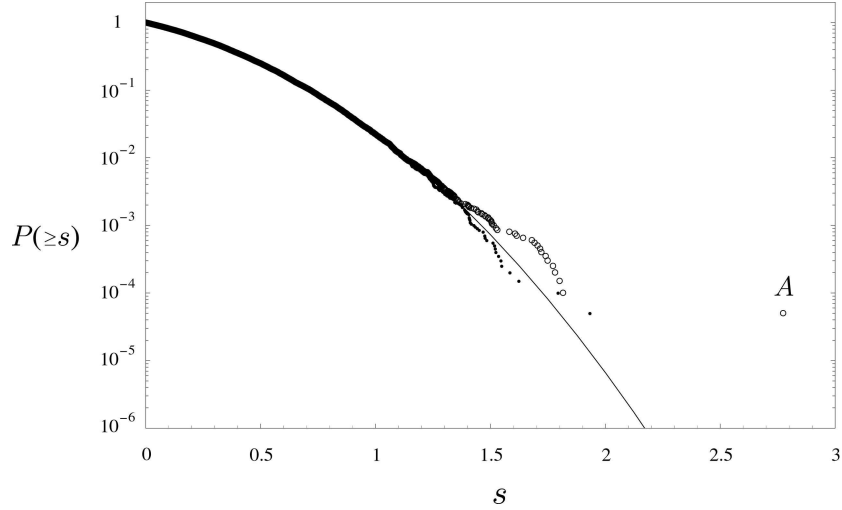}
\caption{The comparison between the Gaussian accumulated distribution
(\ref{eqn:gac}) and the behaviours of the
actual Fourier coefficients of {\it Quarantine}.}
\label{fig:qfgauss}
\end{center}
\end{figure}
It is apparent that the real part of $\tilde{S}_m$
has a significant peak denoted by the letter $A$
at the middle-point $m=K/2$, where $\tilde{S}_{K/2}=-2.662$.
No other structure is visible and the rest of the Fourier coeficients
are consistent with the white noise characteristics (\ref{eqn:whitenoise1}) 
and (\ref{eqn:whitenoise2}).
In fact, the average number of syllables in this section is
$s=1.303$, which leads to $\sqrt{\Delta}=0.628$.
This humber is consistent with the plotted result, as is apparent in Fig.\ref{fig:qfgauss}.
If the random segmentation is valid,
the central limiting theorem dictates that
the distribution of the Fourier coefficients $\tilde{S}_m$
for $m\ne0$ should follow the Gaussian distribution,
\begin{equation}
P_{\rm G}(s)=
\frac1{\sqrt{2\pi}\,\Delta}\exp\left[-\frac{s^2}{2\Delta}\right],
\end{equation}
where $s={\rm Re}(\tilde{S}_m), {\rm Im}(\tilde{S}_m)$.
In Fig.\ref{fig:qfgauss}, the solid curve is the accumulated distribution,
\begin{equation}
P(\ge s)\equiv \int_s^\infty P_{\rm G}(s)ds,
\label{eqn:gac}
\end{equation}
with the above value of $\Delta$.
The dots are plots of the positive ${\rm Re}(\tilde{S}_m)$
versus its rank (the largest being 1, next 2, and so on) devided by
the total number of the positive ${\rm Re}(\tilde{S}_m)$.
The open circles are similar plots for the negative ${\rm Re}(\tilde{S}_m)$.
Within statistical accuracy the dots and open circles should follow
the solid curve if the Gaussian distribution applies.
In Fig.\ref{fig:qfgauss} we see that almost all the points are close to the solid curve, except for the isolated point $A$ at the 
far right, which is the peak $A$ at $m=K/2$ in Fig.\ref{fig:qfourier}.
The fact that this point $A$ is far above the theoretical curve 
suggests that it is very unlikely that this value of $A$ 
is achieved simply by a statistical accident.
This can be explained as follows.
At this point, $s=2.662$, we find that the accumulated probability
distribution is $P(\ge s)=2.037\times10^{-9}$, which means that
the probability that a point exists beyond this value of $s$ is
$2.037\times10^{-9}$. On the other hand, there are 19,861
negative ${\rm Re}(\tilde{S}_m)$ and the point $A$ exist at this $s$.
So the ``measured probability" is $1/19,861=5.035\times10^{-5}$, which
is the vertical coordinate of this point.
Therefore, the existence of this point $A$ simply by accident
is quite unlikely (by the probability $4.045\times10^{-5}$).
Putting this in another way, we may argue as follows:
Since we have $k=K/2$ number of points, the expected number of points
beyond this value of $s$ is equal to $P(\ge s)\times (K/2)=
4.07\times10^{-5}$.  This should be contrasted with the 
existence of a single point $A$, which leads to a conclusion that
the existence of $A$ by chance is almost impossible.
This argument establishes that the peak $A$ in Fig.\ref{fig:qfourier}
is not an accident of statistical fluctuation, but is a real effect.

The existence of the peak $A$ shows that there is a strong period of 
2 in {\it Quarantine}.  This is readily seen from the fact that
the contribution of
the $m=K/2$ term in the expression of the Fourier series (\ref{fouriercom})
is $(-1)^m\tilde{S}_{K/2}$.
Since $\tilde{S}_{K/2}$ is negative, we find that throughout 
{\it Quarantine}, words in even positions (2nd, 4th, 6th, $\cdots$ words)
tend to be shorter than those in odd positions. That is to say, there is a tendency for shorter and longer words to alternate in Crace's text. The relationship between this fact and the rhythmical patterning and consequent lineation noted below may be easily guessed at, but has yet to be examined in any scrupulous way.

\subsection{The Correlation Function}
\begin{figure}[ht]
\begin{center}
\includegraphics[width=11cm]{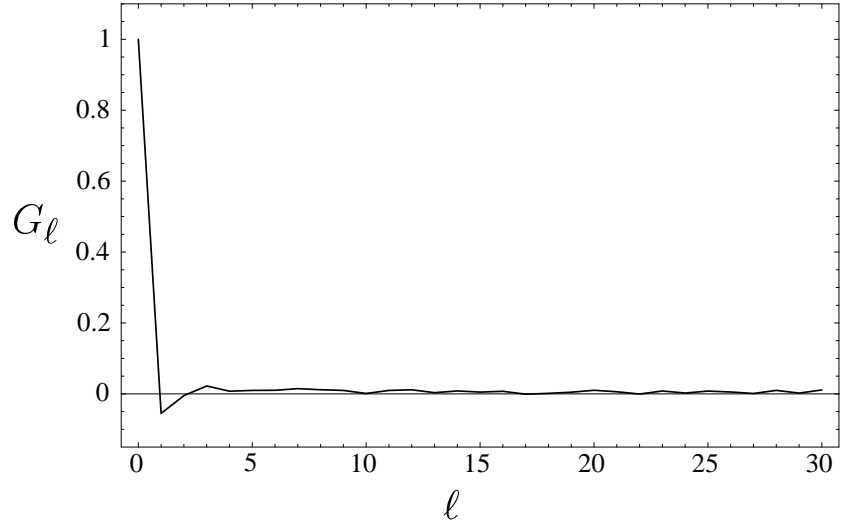}
\caption{The correlation function $G_\ell$ for {\it Quarantine}.}
\label{fig:qcorr}
\end{center}
\end{figure}

The result is plotted in Fig.\ref{fig:qcorr}, where it will be seen 
that there is no apparent structure and the result is 
consistent with the prediction (\ref{deltakk}) from the random segmentation
hypothesis. 
The only deviation from (\ref{deltakk}) is the slight deviation
at $\ell=2$.  We find that this is qualitatively consistent with the analysis
done for the $Q_2$ dip in English prose by Aoyama and Constable (1999).

\subsection{The $Q_n$ distribution}
\begin{figure}[ht]
\begin{center}
\includegraphics[width=11cm]{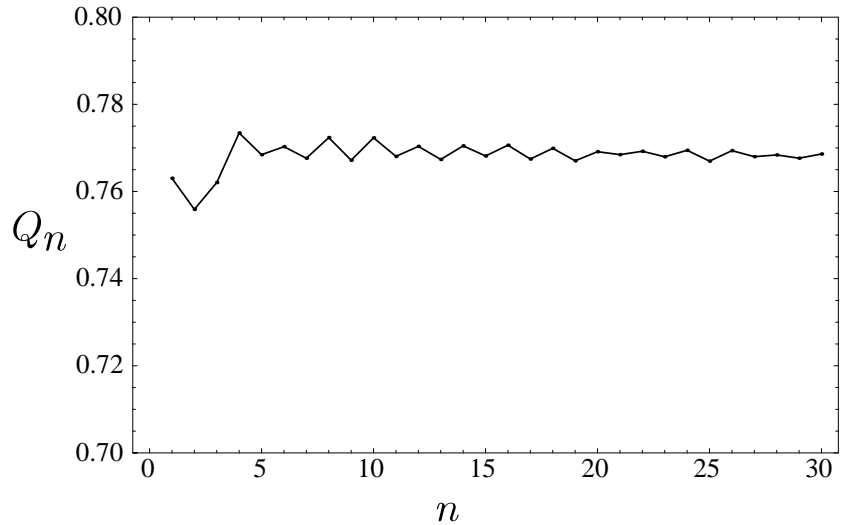}
\caption{The $Q_n$ distribution for {\it Quarantine}.}
\label{fig:qqn}
\end{center}
\end{figure}
Fig.\ref{fig:qqn} shows the $Q_n$ distribution for {\it Quarantine}. The peaks at $n=4$, 6, 8, 10, and so on, are sufficiently evident. It should also be noted that the value for $n=2$ is not, as might be thought, inconsistent with these results. In fact a dip at $n=2$ is normal in English prose (see Aoyama and Constable 1999), and that found in {\it Quarantine} proves to be rather smaller than would otherwise be expected. 

\begin{figure}[ht]
\begin{center}
\includegraphics[width=13cm]{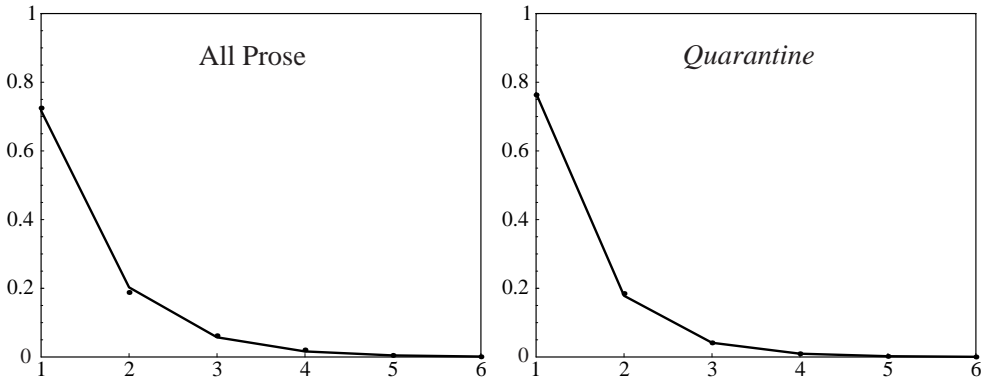}
\caption{The probability distributions of word length (dots) and the 
geometric distribution (solid line) for
all of the prose data in Aoyama and Constable (1999) and {\it Quarantine}.
The geometric distributions were defined so that the average number
of syllables per word agrees with the actual distributions for 
each case.}
\label{fig:pq}
\end{center}
\end{figure}

In other words what appears to be a puzzling dip at $n=2$, where one might naively expect a peak relating to those at multiples of two, is in fact a smaller than usual depression. 
The underlying causes of this reduced depression appear to be an enhancement of disyllables, as can be seen by comparing the word length distribution of Crace's text with both an ideal geometric distribution and the data from a large sample of English prose (See Fig.\ref{fig:pq}).
Crace uses slightly more disyllables than would be predicted from the geometric distribution, whereas the prose corpus texts employ slightly less.

\section{T. S. Eliot \it Four Quartets}
\subsection{Fourier coefficients}
We have found that the Fourier coefficients $\tilde{S}_m$
for any of the four sections
have no significant structure, unlike {\it Quarantine}.
They are all consistent with 
the white noise characteristics (\ref{eqn:whitenoise1}) 
and (\ref{eqn:whitenoise2}).

\subsection{The Correlation Function}
\begin{figure}[ht]
\begin{center}
\includegraphics[width=11cm]{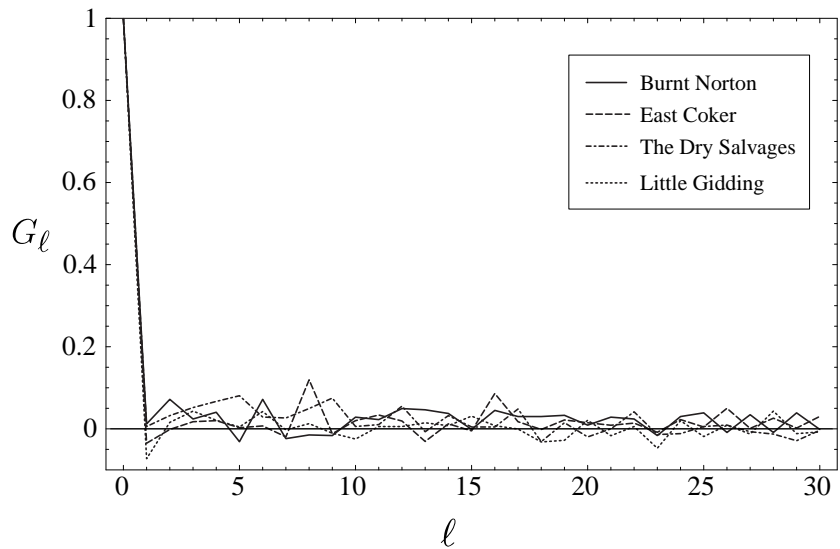}
\caption{The correlation function $G_\ell$ for each of the sections.}
\label{fig:fcorr}
\end{center}
\end{figure}

The result is plotted in Fig. \ref{fig:fcorr}, where it will be seen 
that there is no apparent structure.  

\begin{figure}[ht]
\begin{center}
\includegraphics[width=11cm]{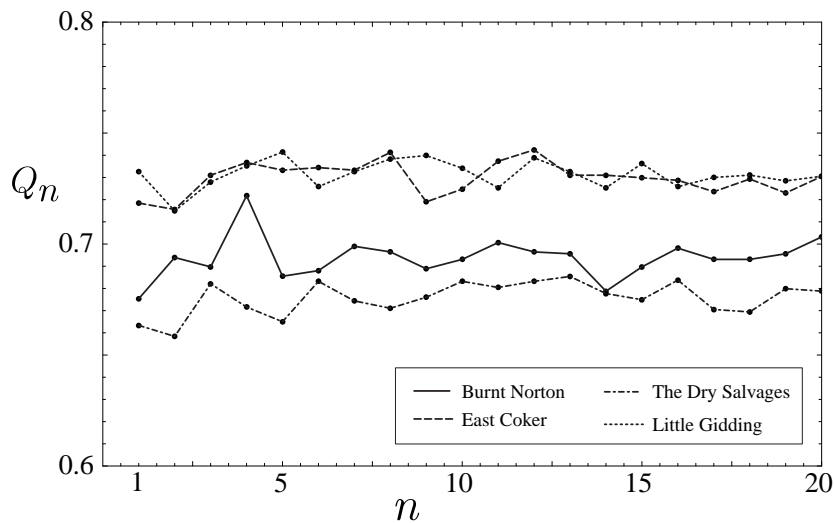}
\caption{The $Q_n$ plot of the four sections.}
\label{fig:fqn}
\end{center}
\end{figure}

\subsection{The $Q_n$ distribution}
Fig.\ref{fig:fqn} shows these $Q_n$ distributions 
for each of all four sections of {\it The Four Quartets}.
As is apparent in this plot,  {\it Burnt Norton} alone
has a marked peak,  at $n=4$, a fact which suggests that a substantial 
part of it is composed in units of four syllables.\footnote{Similar peaks
are observed for isometrically lineated verse. See Constable
and Aoyama (1999).}
Of course due to the randomness of the original syllable distribution,
one might observe a similar peak by chance. Some care is required in
determining whether this $Q_4$ peak is the result of authorial 
compositional ordering, or a simple accident.
We can determine the statistical significance of the peak $Q_4$ 
by calculating the average 
($\bar Q$) and the standard deviation ($\sigma_Q$)
of $Q_n$ from $n=1$ to 200 . For the {\it Burnt Norton} section 
we find the following values:
\begin{eqnarray}
\bar{Q}&\equiv&\frac1N \sum_{n=1}^N Q_n=0.6914,\cr
\sigma_Q&\equiv&\frac1N \sqrt{\sum_{n=1}^N (Q_n-\bar{Q})^2}=0.007219,
\end{eqnarray}
where $N=200$.  
The value of the $Q_4$ in this section is,
\begin{equation}
Q_4=0.7218\simeq\bar{Q}+4.2\,\sigma_Q,
\end{equation}
showing that this is a ``4$\sigma$ effect".

\begin{figure}[ht]
\begin{center}
\includegraphics[width=11cm]{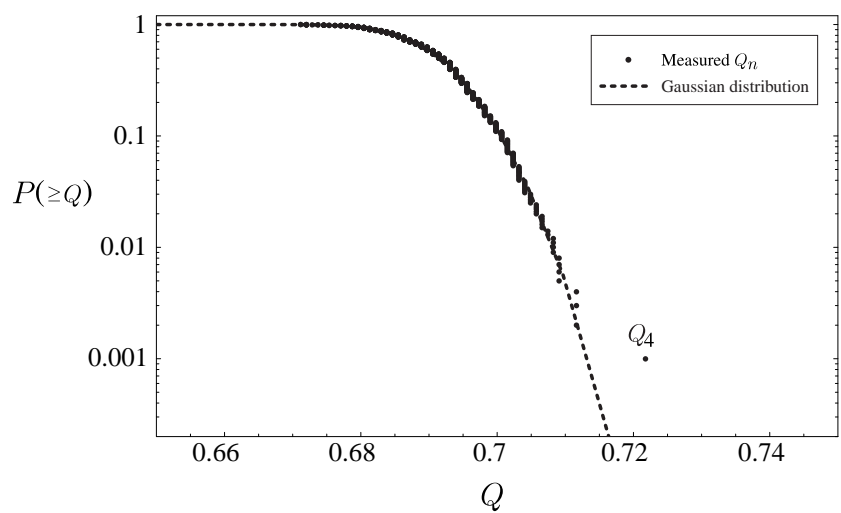}
\caption{The log-linear plot of the accumulated probability density
$P(\ge Q)$ (dashed line) and the distribution of the
measured $Q_n$ $(n=1\sim1000)$ for the {\it Burnt Norton} section.
The agreement with the Gaussian distribution is evident, 
except for $Q_4$.}
\label{fig:gauss}
\end{center}
\end{figure}
Another way to guage the significance of the $Q_4$ peak is to make a plot similar to Fig.\ref{fig:qfgauss}.
The probability of such a large deviation from the mean
value $\bar{Q}$ is given by the accumulated
probability $P(\ge Q)$, which is defined as the probability
that a value larger than or equal to $Q$ is observed;
\begin{equation}
P(\ge Q)\equiv\int_Q^\infty P_{\rm G}(Q')\,dQ',
\label{eqn:gauss}
\end{equation}
where $P_{\rm G}(Q)$ is the Gaussian distribution;
\begin{equation}
P_{\rm G}(Q)=\frac1{\sqrt{2\pi}\,\sigma_Q}
\exp\left[-\frac{(Q-\bar{Q})^2}{2\sigma_Q^2}\right].
\end{equation} 
For $Q_4=0.7218$, we find that $P(\ge Q_4)=1.3\times 10^{-5}$.
Straightforwardly, the probability of observing as large a peak 
as that for $Q_4$ simply by accident is about $1.3\times 10^{-5}$.
This situation is shown graphically in Fig.\ref{fig:gauss},
which is similar to Fig.\ref{fig:qfgauss}.
In this figure, the accumulated probability density
$P(\ge Q)$ defined in Eq.(\ref{eqn:gauss}) is shown by the dashed line
and the actual distribution of $Q_n$ in this section by dots.
The agreement between the continuous Gaussian distribution
with the actual measured value is excellent, {\it except for $Q_4$}.
This both justifies the use of the Gaussian distribution above, 
and visualizes the abnormality of the $Q_4$ peak.

This result is consistent with the lack of any structure in
the Fourier coefficients $\tilde{S}_n$, since the latter are
only sensitive to periodic structures involving the syllable numbers
$\{S_1, S_2, \cdots, S_K\}$, and would be ideal for studying 
such structures. That is to say, if the four syllable units
appear in combination and in a periodic manner, 
the Fourier coefficients would make their existence evident. However, 
four syllable units, may appear in various combinations,
of 1 and 3 syllables, 2+2, 1+1+2, and so on; furthermore, they may appear randomly in the section.
Both of these facts would disrupt the periodic structure, and
render the Fourier analysis useless.
The same is true for the correlation function $G_\ell$.
On the other hand, our word-boundary correlation function $Q_n$ is by definition
sensitive to the existence of the four syllable units
even in this situation. 

\section{Conclusion and Further Comments}

We have shown that the $Q_n$ computation is a sound and fairly sensitive register of one fundamental feature of lineated text, and complementary in some respects to two other statistical methods. In concluding what has been so far a methodological paper we will return to the language materials under consideration and offer several brief comments on the significance of the $Q_n$ peaks detected in the {\it Four Quartets} and {\it Quarantine}. 

\subsection{T. S. Eliot, {\it Four Quartets}}
The metrical status of Eliot's {\it Four Quartets} has been much debated, notably in Cooper (1998). $Q_n$ analysis will not resolve all aspects of this debate, but  does contribute significantly in regard to lineation. Our findings are that with the exception of {\it Burnt Norton} the {\it Quartets} are not mathematically lineated. In this we are in basic agreement with Cooper, who carried out an exhaustive rhythmical scansion. We supplement his remarks by observing that whilst the later three {\it Quartets} appear on reading to be to some degree rhythmically regular (and there are passages which are very obviously composed in isometric lines), {\it Burnt Norton} alone is mathematically lineated overall, the basic mathematical line being a segment of four syllables. It seems likely that this results from a two beat duple segment, running as follows: offbeat, beat, offbeat, beat. It should be noted from the $Q_n$ distribution that there is no subsequent peak at {\it n} = 8, suggesting that while these four syllable segments are frequent, they are not so frequently adjacent as to produce an eight syllable mathematical line. This is surprising, and suggests deliberate avoidance on Eliot's part, as if he were unwilling to let his rhythms move too close to the familiar four beat octosyllabic unit.

It seems possible that the poem as it survives traces an experiment with rhythmical patterning, and that Eliot's neglect of the 4 syllable line in the subsequent sections is a refinement of method, and the adoption of a revised compositional principle which permitted the production of the desired rhythmical effects without relying on clear echos of regular metrical structures. {\it Burnt Norton} is an early attempt to employ rhythmical metre in a form which is looser in its lineation structure than standard verse. As it happens it is still mathematically lineated. The subsequent three Quartets reveal further and more adventurous developments of this experiment. It would be interesting to know whether this anticipates in any way the character of the metre employed in Eliot's later plays, and we note this as being a topic for future research.

\subsection{Jim Crace, {\it Quarantine}}
Crace's writing, certainly the more recent books, has been very widely regarded as of a type differing from standard prose. The critic Frank Kermode has noted of {\it Quarantine} that it is from `the end of the fiction spectrum where the novel is most like a poem, most turned in on itself, most closely wrought for the sake of art and internal cohesion' (Kermode 1998). John Banville, discussing another of Crace's novels, {\it Being Dead}, has even pointed out that technically the prose often uses verse fragments, and that much of this book is `written in a kind of broken blank verse, and indeed could be successfully laid out in verse form', adding that Crace is `particularly fond of iambic pentameter' 
(Banville 2000). 

Certainly, {\it Quarantine} is readily recognized as being rhythmically more regular than ordinary prose, but there are no visually salient lines, and even careful examination during reading fails to find any convincing or consistent division. Although it is possible, as Banville says of {\it Being Dead}, to find occasional lines, even short sequences, it is difficult to successfully display any sizeable piece of the text as isometric verse. Casual examination can make no further progress. However, the $Q_n$ procedure is sufficiently sensitive to detect lineation even when distributed in small and isolated packets, and of {\it Quarantine} it reveals that syllabic groups of two, four, six (which is, curiously, less prominent), eight, ten, and subsequent multiples of two are significantly more common than they would be in standard prose. {\it Quarantine} is non-randomly segmented, and even though it does not employ a core isometric line length, and its `lines' do not follow on one from another, it is still, and in a novel and important sense, {\it lineated}.

The fact that there is no particular preference for one line length over another, that is no preference, say, for decasyllabic groups over octosyllabic groups, is of very great interest, and suggests that Crace did not set out to write in lines and then to conceal them by printing his work with a prose layout. Instead we suspect that the lineation detected by the $Q_n$ procedure arises as a byproduct of a deliberate organisation of the rhythmical patterning of the text.

The character of this general organisation can be inferred from the $Q_2$ data. As noted above (see Section 3.3), although apparently anomalous, because it seems to register a depression rather than a peak, this data point is in fact consistent with the subsequent peaks at multiples of two, since the depression in Crace's text is actually smaller than expected. The fundamental cause of this, as noted in our discussion above, is simply that Crace has employed slightly more disyllabic words than would be expected in standard prose, and thus we may conclude that lineation arises in Crace's text from a predisposition to employ disyllabic words so as to facilitate the construction of alternating rhythms.

As far as we know Crace has not commented in public on his reasons for employing this structuring. If asked he would perhaps reply, taking up terms similar to those used by Kermode above, that he felt it gave his work a self-supporting and self-consciously artistic structural brace that unpatterned prose did not possess. While this might be a satisfactory proximal psychological explanation, we suspect that a deeper account of the resulting experiences for both reader and writer is available and may more adequately account for the appeal of this formal device. Anyone who has read Crace's work will agree with John Updike that it has a `hallucinatory' quality (Updike 1999). That this is not readily accounted for by reference to the facts of the narrative is worth remark, and is  a quality which recalls the character of much of the most successful poetry.

The exact consequences of the variety of mathematical lineation found in {\it Quarantine} are yet to be fully understood, but it is so far clear that at least some of the previous restrictions noted in regard to isometrically lineated text also apply to Crace's composition. The peaked $Q_n$ distribution indicates that some degree of syntactic distortion occurs, and we have seen above that dictional distortion is present in the enhancement of disyllabic words. It is also possible that even though he was not composing deliberately in a single isometrical line, but varying his segment length, that some reduction in mean word length has taken place, in order to increase the syntactical options (a known consequence of isometric lineation: see Constable and Aoyama 1999). Both these distortions have been discussed by Constable (1998) within his Disruption Theory, which proposes that readers attempting to process text which is evenly and subtly disrupted can neither reject it as damaged nor successfully interpret it, and consequently experience an illusion of profundity (see also the sketch in Constable and Aoyama (1999)). However, it seems almost certain that with the variety of lineation employed in Crace's {\it Quarantine} these disruptions are less marked. However, it should be noted that a further source of disruption, that caused by rhythmical patterning, may well be present, namely an alteration in the frequency of stressed and unstressed syllables leading to an increase in the frequency of content terms. This disruption is also found in standard duple verse, but in Crace's text it may be the principal source of disruption, with syntactical and word length distortions playing a much smaller part. The tendency to use more disyllables is relevant here, though its dictional effects, on the frequency of the various parts of speech for example, are at present unknown.

On such a view the text is experimental in the sense that it varies the proportions of the disruptive influences in order to produce new experiences for readers. It might be seen as an attempt to provide prose with some of the richness of effect familiar from isometric verse without suffering to the same degree its disablements, and should be compared with experimental works such as Eliot's {\it Four Quartets} which approach the problem from the other side, and beginning from a verse base move towards the freedom of prose without losing the benefits of disruption (see Constable (1998) for a discussion of the decline of isometric verse and the rise of prose in terms of disruption theory).

\subsection{The Roots of Lineation and Formal Innovation}
Comparison of these two texts sheds light on the relationship between mathematical lineation and rhythmical organisation. In one case, {\it Four Quartets}, we have rhythmical organisation with and without mathematical lineation, whereas in {\it Quarantine} we find both mathematical lineation and rhythmical organisation, but with lineation only appearing as an unconscious byproduct of the attempt to create regular rhythmical effects.

This may shed some light on the roots of visible and salient lineation as a metrical rule, since we can see that even when an author is not aware of lineation as a constraint, which is arguably the case in preliterate composers, lineation may arise as a byproduct of rhythmical regularity. What we have in Crace is some evidence suggesting that lineation is a likely, indeed a very probable outcome of rhythmical organisation. We hypothesize that the roots of lineation as a metrical restriction are deep, at least in English, and that mathematical lineation may be coeval with rhythmical patterning. Its extraction and development into a separately understood metrical object, the line, is likely to have come much later. What is particularly interesting is that while lineation should have been recognized as independent of rhythm (i.e. in syllabic verse), there has been no parallel recognition of regular rhythmical patterning without isometric lineation.

But, and this is the most interesting point arising from Eliot's texts, lineation would appear to be unnecessary to some degree of rhythmical organisation, and with care it seems possible that it may be avoided. We speculate that Eliot's experimental work in {\it Four Quartets} is not an isolated example, and that others may be found in Seventeenth Century Drama, particularly in Shakespeare. Nevertheless, even if such antecedents are found it seems likely that regularly rhythmical but unlineated text is a largely unexplored technical possibility, one which does not come about by chance with any great frequency and one which is difficult to achieve deliberately, certainly without detailed knowledge. It is also possible that given other features of English, namely those relating to word length, part of speech frequency, and stress, that mathematical lineation is an all but unavoidable consequence of definite rhythmical patterning if noticeably unnatural output is to be avoided. Further theoretical work is expected to permit more conclusive comments on this matter.

Interesting though {\it Quarantine} is, our tests show that there is still room for further experimentation in this vein, and the abstract understanding of lineation offered in this paper and our previous papers, may go some way to assisting writers in devising composition methods which further minimize lineation. The psychological effects of text of this type are quite unknown, but it is conceivable that a diction somewhat distorted by the construction of rhythmical regularity would combine with very well with a syntax undisrupted by non-random word length ordering, and perhaps produce a formal constraint as vigorous and attractive to readers as isometric verse once was.

\subsection{The Prose-Verse Continuum}
The characters of Eliot's {\it Quartets} and Crace's {\it Quarantine} as revealed here both act as an incitement to reconceive the relationship between verse and prose. Colloquial and academic literary critical discussion, has generally treated `verse' and `prose' as intuitively obvious exclusive categories separated by an ill-defined grey area. This essentialist tendency has been encouraged by the binary leanings of generative metrics, and our own work on the computable distinction between prose and verse might even be seen as reinforcing this habit. However, there is nothing in the facts of the matter which obliges us to conceive of prose and verse in this rigid way, and we believe that careful consideration of our theoretical research will show that there are very good reasons for taking up an alternative conception, namely that what we think of as {\it typical verse} and {\it typical prose} are {\it distinguishable clusters of texts positioned on a continuum}.

Underlying such a view is the fundamental insight that while isometric verse involves an ordering of word length this is not a feature which is either absolutely present or absent, but one that varies in degree between texts. Thus we may say that there is a continuum between random disorder of word length on one side and maximal order on the other. From an abstract theoretical perspective we can see that any particular text may lie at some point along this continuum, but practically speaking there are considerable difficulties in the way of assigning it to a precise position. However, this need not deter us from recognizing the utility of the continuum as a conception, and in this light the difficulties of text assignment appear as technical challenges rather than as obstacles to accepting the continuum in the first place.

A key issue in addressing this matter will be a refined understanding of word length ordering. All the prose texts we have so far studied in fact exhibit a very small degree of order in the fine structure of the word length profile (see Aoyama and Constable (1999), and Constable and Aoyama (1999)), and so seem to lie at some distance from the theoretical prose pole, while at the other extreme no text so far examined is without randomness in its word length characteristics, and thus lies at some distance from the theoretical verse pole. In between we can be fairly confident that some texts we have studied lie closer to the theoretical maximum of order, such as Longfellow's {\it Hiawatha}, which has a very marked repeating structure in its $Q_n$, while others seem to lie closer to the disordered pole, such as most blank verse by Wordsworth, whose normative line length peaks are modest in size and whose multiple peaks diminish rapidly as a result of weak long range correlation (i.e. high frequency of variant lines). The location of a text such as Crace's {\it Quarantine} is more difficult to determine. As noted above, while it exhibits a repeating structure it has no major peak (i.e. no predominant line length), but instead a whole series of minor peaks. At present we are not in a position to distinguish definitively between the degree of order in such a text and that in, for example, a fundamentally isometric text such as Wordsworth's {\it Prelude}. It is tempting to suppose that {\it Quarantine} must lie closer to the prose end of the continuum, on account of its lack of concentrated order, but this, though convenient, may be no more than a residual effect of the exclusive categories that the conception of the continuum was introduced to replace. Satisfactory resolution of this matter will not be possible until some better method has been determined for remarking upon the degrees of order and disorder in a text's lineation structure, and it may be noted here in passing that the Fourier analysis will probably of use in this regard.


\end{document}